\documentclass{article} 
\usepackage{iclr2026_conference,times}

\usepackage{amsmath,amsfonts,bm}

\def\eqref#1{equation~\ref{#1}}

\def\1{\bm{1}}

\DeclareMathAlphabet{\mathsfit}{\encodingdefault}{\sfdefault}{m}{sl}
\SetMathAlphabet{\mathsfit}{bold}{\encodingdefault}{\sfdefault}{bx}{n}

\usepackage[table]{xcolor}
\usepackage{url}
\usepackage{booktabs}
\usepackage{graphicx}
\usepackage{subcaption}
\usepackage{float}
\usepackage{etoc}
\usepackage{multirow}
\usepackage{wrapfig}
\usepackage{tcolorbox}
\usepackage{xcolor}
\usepackage[colorlinks=true, urlcolor=blue, linkcolor=blue, citecolor=blue]{hyperref}
\usepackage{comment}
\tcbuselibrary{breakable,listings}
\usepackage{amssymb}

\usepackage{xcolor}
\usepackage{makecell}
\usepackage{booktabs}
\newcommand{\good}{\textcolor{green!60!black}{\checkmark}}
\newcommand{\bad}{\textcolor{red!70!black}{\textsf{x}}}
\newcommand{\semi}{\textcolor{yellow!70!black}{\textsf{•}}}

\usepackage{tikz}
\usepackage{pgfplots}
\usepgfplotslibrary{polar}
\pgfplotsset{compat=1.18}

\title{HealthAdminBench: Evaluating Computer-Use Agents on Healthcare Administration Tasks}

\author{
\normalfont
Suhana Bedi$^{1,*}$ \quad
Ryan Welch$^{1,*}$ \quad
Ethan Steinberg$^{2,*}$ \quad
Michael Wornow$^{2}$ \quad
Taeil Matthew Kim$^{1}$ \\
Haroun Ahmed$^{2}$ \quad
Peter Sterling$^{2}$ \quad
Bravim Purohit$^{2}$ \quad
Qurat Akram$^{3}$ \quad
Angelic Acosta$^{3}$ \\
Esther Nubla$^{3}$ \quad
Pritika Sharma$^{3}$ \quad
Michael A. Pfeffer$^{1,3}$ \quad
Sanmi Koyejo$^{1}$ \quad
Nigam H. Shah$^{1,3}$ \\
\\
$^{1}$ Stanford University \\
$^{2}$ Kinetic Systems \\
$^{3}$ Stanford Health Care \\
$^{*}$ These authors contributed equally to this work.
}
\iclrfinalcopy 
\begin{document}

\maketitle
\begin{abstract}
Healthcare administration accounts for over \$1 trillion in annual spending, making it a promising target for LLM-based computer-use agents (CUAs). While clinical applications of LLMs have received significant attention, no benchmark exists for evaluating CUAs on end-to-end administrative workflows. To address this gap, we introduce \textsc{HealthAdminBench}, a benchmark comprising four realistic GUI environments—an EHR, two payer portals, and a fax system—and 135 expert-defined tasks spanning three administrative task types: Prior Authorization, Appeals and Denials Management, and Durable Medical Equipment (DME) Order Processing. Each task is decomposed into fine-grained, verifiable subtasks, yielding 1,698 evaluation points. We evaluate seven agent configurations under multiple prompting and observation settings and find that, despite strong subtask performance, end-to-end reliability remains low: the best-performing agent (Claude Opus 4.6 CUA) achieves only 36.3\% task success, while GPT-5.4 CUA attains the highest subtask success rate (82.8\%). These results reveal a substantial gap between current agent capabilities and the demands of real-world administrative workflows. \textsc{HealthAdminBench} provides a rigorous foundation for evaluating progress toward safe and reliable automation of healthcare administrative workflows. We release the benchmark, environments, and leaderboard at \url{https://healthadminbench.stanford.edu}.
\end{abstract}

\section{Introduction}

Large language models (LLMs) in healthcare have primarily been evaluated on clinical tasks such as diagnostic reasoning and triage \citep{arora2025healthbench, bedi2024testing}. In contrast, administrative tasks, including prior authorization, claims submission, and patient intake, remain underexplored despite accounting for over \$1 trillion in annual U.S. healthcare spending \citep{sahni2023administrative} and contributing to preventable medical errors \citep{makary2016medical, statpearls2024error}. These tasks require multi-step interactions across legacy enterprise systems that lack API support, limiting direct programmatic automation. 

LLM-based computer-use agents (CUAs), which observe and interact with software interfaces, offer a potential solution. However, existing benchmarks do not capture the multi-step and cross-system complexity of administrative healthcare tasks. Prior work focuses on web navigation tasks on public websites \citep{zhou2023webarena, deng2023mind2web, garg2025realbenchmarkingautonomousagents, xu2025webbenchllmcodebenchmark} and general enterprise software \citep{drouin2024workarenacapablewebagents}, and does not reflect the constraints of healthcare systems, which involve gated access, limited documentation, and sensitive data. Healthcare-specific efforts partially address this gap. MedAgentBench \citep{jiang2025medagentbench} provides an interactive environment for LLM-based agents, but is restricted to clinical tasks implemented via Fast Healthcare Interoperability Resources (FHIR) API calls and does not support GUI-based interaction. MedHELM \citep{bedi2026medhelm} includes administrative tasks within a broader evaluation framework, but is limited to static, text-based evaluation and does not capture the dynamic, multi-step environments required for computer-use agents.

To address this gap we introduce \textbf{\textsc{HealthAdminBench}}, a benchmark for evaluating computer-use agents in healthcare administration. We design multi-step, GUI-based tasks derived from real-world administrative tasks observed through hospital shadowing, and use the “Administration and Workflow” category from MedHELM to guide task selection. We construct four deterministic web environments simulating core administrative systems, including an electronic health record (EHR), two payer portals, and a fax system. Within these environments, we define 135 expert-designed tasks grouped into three administrative task types corresponding to common revenue-cycle workflows: Prior Authorization, Appeals and Denials Management, and Durable Medical Equipment (DME) Order Processing

Building on the REAL framework \citep{garg2025realbenchmarkingautonomousagents}, we define verifiers (1,177 deterministic and 521 LLM-based) for 1,698 subtasks, enabling fine-grained evaluation of agent behavior. We evaluate seven agent configurations derived from five frontier models (including both our harness and native computer-use agents) and find that the strongest agent achieves a 36.3\% full-task success rate. Although agents perform well on individual subtasks, they fail to reliably complete end-to-end tasks due to errors in document handling and cross-system data transfer.

Our contributions are as follows:
\vspace{-0.5em}
\begin{enumerate}
    \item \textbf{Interactive environments for healthcare administration.} We publish four deterministic computer-use environments to replicate common administrative interfaces, including an Electronic Health Record (EHR), two payer authorization portals, and a fax portal. These environments serve as reproducible proxies for proprietary administrative systems that are not publicly accessible, and can be used for both training and evaluation.

    \item \textbf{Benchmark of 135 specialized administrative tasks.} We curate 135 expert-designed, multi-step tasks derived from real-world administrative work and aligned with the MedHELM "Administration and Workflow" category. Each task is decomposed into verifiable subtasks using a combination of deterministic checks and LLM-based judges, enabling fine-grained evaluation and reward construction.

    \item \textbf{The first evaluation of computer-use agent reliability in healthcare administration.} We evaluate seven agent configurations across five frontier models from multiple providers on \textsc{HealthAdminBench} and find that the strongest agent achieves only a 36.3\% full-task success rate. Although agents often succeed on individual subtasks, they struggle to reliably complete end-to-end tasks, with failures frequently arising from document handling and cross-system coordination. We further show that fine-tuning an open-source model on just 100 training tasks yields a 23\% absolute improvement in held-out task success, surpassing the best closed-source agent by 14\%. These results suggest that training on these interactive environments may substantially improve agent performance.
\end{enumerate}

To support reproducible benchmarking, we release all environments, task definitions, evaluation code, and model outputs. The public leaderboard is hosted at \url{https://healthadminbench.stanford.edu}, while the codebase is available at \url{https://github.com/som-shahlab/health-admin-bench}

\begin{figure}[!h]
    \centering
    \includegraphics[width=\linewidth]{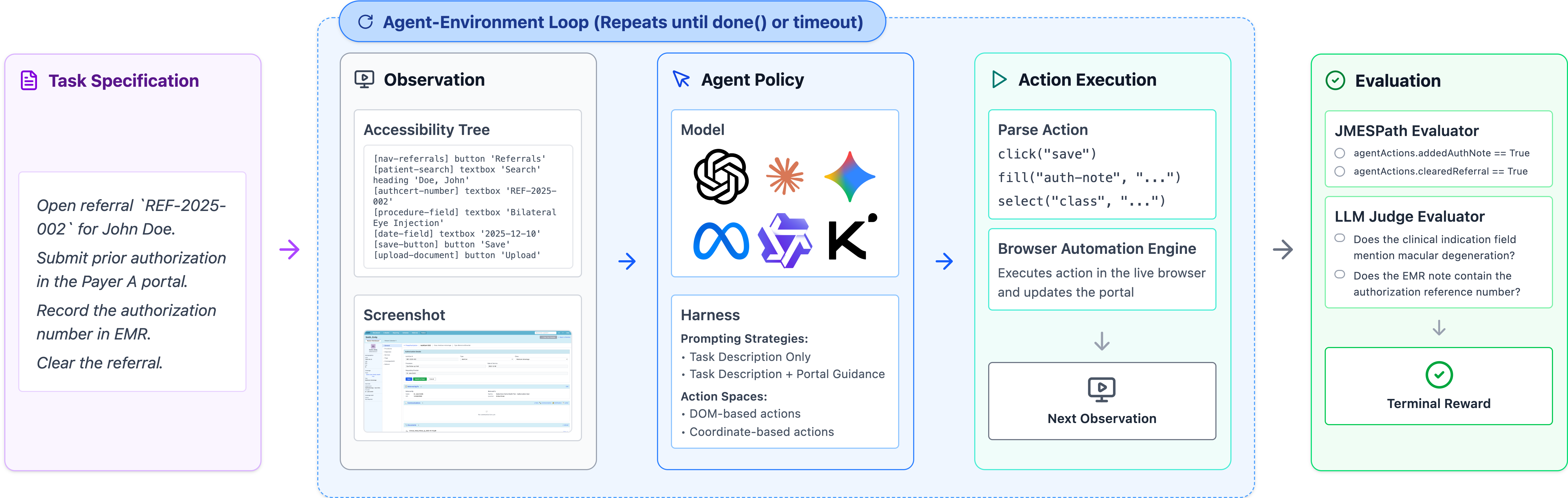}
    \caption{\textsc{HealthAdminBench} evaluation loop. Each task is executed by an agent through iterative observation, action selection, and interaction with simulated environments (EHR, payer portals, and fax). Task success is determined using a combination of deterministic checks and LLM-based judges.}
    \label{fig:main}
\end{figure}

\section{Background and Related Work}

\vspace{-0.5em}
\subsection{Computer-Use Agents}

\paragraph{Web interaction.}
Prior work on web agents spans synthetic environments such as \textsc{MiniWoB++} \citep{liu2018reinforcement}, which focus on basic UI primitives, to more realistic benchmarks such as \textsc{WebShop}, \textsc{WebArena}, and \textsc{Mind2Web} that introduce longer-horizon, multi-website tasks \citep{yao2023webshopscalablerealworldweb, zhou2023webarena, deng2023mind2web}. More recent work incorporates multimodal interaction via screenshots (\textsc{WebVoyager}) and deterministic replicas of real-world environments with programmatic evaluation (\textsc{REAL}) \citep{he2024webvoyagerbuildingendtoendweb, garg2025realbenchmarkingautonomousagents}.

\vspace{-0.5em}
\paragraph{Enterprise and ``workplace'' web environments.}
Recent benchmarks shift toward enterprise workflows with structured schemas and multi-step constraints. \textsc{WorkArena} and \textsc{REAL} evaluate agents in realistic settings such as CRM and ticketing systems, while frameworks like BrowserGym unify these environments under a common interface for large-scale evaluation \citep{drouin2024workarenacapablewebagents, garg2025realbenchmarkingautonomousagents, dechezelles2025browsergymecosystemwebagent}. However, these benchmarks focus on general enterprise tasks and do not capture domain-specific workflows such as healthcare administration, which require coordination across multiple systems and organizations.

\vspace{-0.5em}
\paragraph{OS and multi-application workflows.}
Beyond the browser, benchmarks now test full desktop environments. \textsc{OSWorld} \citep{xie2024osworldbenchmarkingmultimodalagents} evaluates multimodal agents on open-ended tasks in real OS environments, requiring GUI grounding, tool switching, and recovery from errors. These results suggest that failures often arise from brittle perception-action grounding and incomplete operational knowledge rather than purely ``reasoning'' mistakes. These challenges are amplified in healthcare administration, where tasks require coordination across multiple environments and reliable execution of schema-constrained interactions.

Table~\ref{tab:benchmark-taxonomy} summarizes how these benchmarks compare along dimensions that matter for administrative work, including cross-application workflows, schema-constrained interactions, explicit commit actions, and objective success checks.

\begin{table}[H]
    \centering
    \footnotesize
    \setlength{\tabcolsep}{4pt}
    \begin{tabular*}{\linewidth}{@{\extracolsep{\fill}} l c c c c}
        \toprule
        \textbf{Benchmark} & 
        \makecell[c]{\textbf{Cross-App} \\ \textbf{Workflows}} & 
        \makecell[c]{\textbf{Strict Forms} \\ \textbf{\& Commits}} & 
        \makecell[c]{\textbf{Objective} \\ \textbf{Success}} & 
        \makecell[c]{\textbf{Healthcare} \\ \textbf{Admin}} \\
        \midrule
        WebShop (2022)             & \bad  & \good & \good & \bad \\
        Mind2Web (2023)            & \bad  & \bad  & \bad  & \bad \\
        WebArena (2023)            & \bad  & \good & \good & \bad \\
        WorkArena (2024)           & \bad  & \good & \good & \bad \\
        OSWorld (2024)          & \good & \good & \good & \bad \\
        REAL (2025)          & \bad & \good & \good & \bad \\
        MedAgentBench (2025)       & \bad  & \good & \good & \bad \\
        MedHELM (2026)       & \bad  & \bad & \good & \semi \\
        \midrule
        \textbf{\textsc{HealthAdminBench}} & \good & \good & \good & \good \\
        \bottomrule
    \end{tabular*}
    \caption{Comparison of \textsc{HealthAdminBench} with related benchmarks. 
    \good = present, \bad = absent. Columns denote: (1) workflows spanning multiple 
    disjoint applications, (2) UI environments with rigid schema validation 
    and terminal commit actions, (3) verifiable, programmatic success criteria, 
    and (4) tasks grounded in healthcare administration.}
    \label{tab:benchmark-taxonomy}
\end{table}

\vspace{-2.0em}
\subsection{Gaps in Evaluating Administrative Healthcare Tasks}

\paragraph{Clinical reasoning and dialogue.}
Benchmarks and evaluation suites such as MultiMedQA \citep{singhal2022largelanguagemodelsencode} and Med-HALT \citep{pal2023medhaltmedicaldomainhallucination} evaluate diagnostic reasoning, clinical question answering, and safety vulnerabilities. More recent comprehensive evaluations such as MedHELM \citep{bedi2026medhelm}, assesses LLMs across a broad range of tasks, including clinical decision support, patient communication, documentation, medical research, and healthcare administration. However, these evaluations are largely text-based and do not capture the end-to-end, payer-facing tasks (e.g., portal navigation, form completion, and submission) that underpin revenue-cycle operations.

\vspace{-0.5em}
\paragraph{Agents in EHR environments.}
Prior benchmarks for interactive healthcare agents have focused mainly on clinical workflows within single, unified EHR environments. MedAgentBench \citep{jiang2025medagentbench, medagentbenchv2} and FHIR-AgentBench \cite{fhiragentbench}, for instance, evaluate agents over structured EHR data for tasks such as lab ordering, note review, and retrieval-based question answering. However, administrative workflows such as prior authorization, appeals and denials management, and DME order processing require coordination across payer portals, external systems, and document exchange, which these benchmarks do not capture. Existing work on automating parts of these processes \cite{pandey2024advancing, vatsal2024gptpriorauth} has also largely emphasized decision support rather than end-to-end execution in realistic environments. \textsc{HealthAdminBench} fills this gap with reproducible, multi-system tasks grounded in payer--provider interactions

\vspace{-1em}
\section{HealthAdminBench}
\label{sec:benchmark}
\vspace{-0.5em}

We define administrative healthcare tasks as sequential decision problems over interactive web environments, formalized as partially observable Markov decision processes (POMDPs). The latent state $s_t \in \mathcal{S}$ captures the full configuration of all browser sessions (e.g., page content, navigation context, and task-relevant statuses), while the agent observes a partial view $o_t = O(s_t)$, consisting of a screenshot or structured interface representation.

At each timestep, the agent selects an action $a_t \in \mathcal{A}$ based on the task specification and interaction history, which is executed via a browser automation engine (Playwright) to deterministically transition the environment to $s_{t+1} = T(s_t, a_t)$. Episodes terminate upon task completion or when a fixed interaction budget is reached, and performance is measured via a binary terminal reward indicating whether the task was successfully completed. 

\begin{figure}
    \centering
    \includegraphics[width=0.9\linewidth]{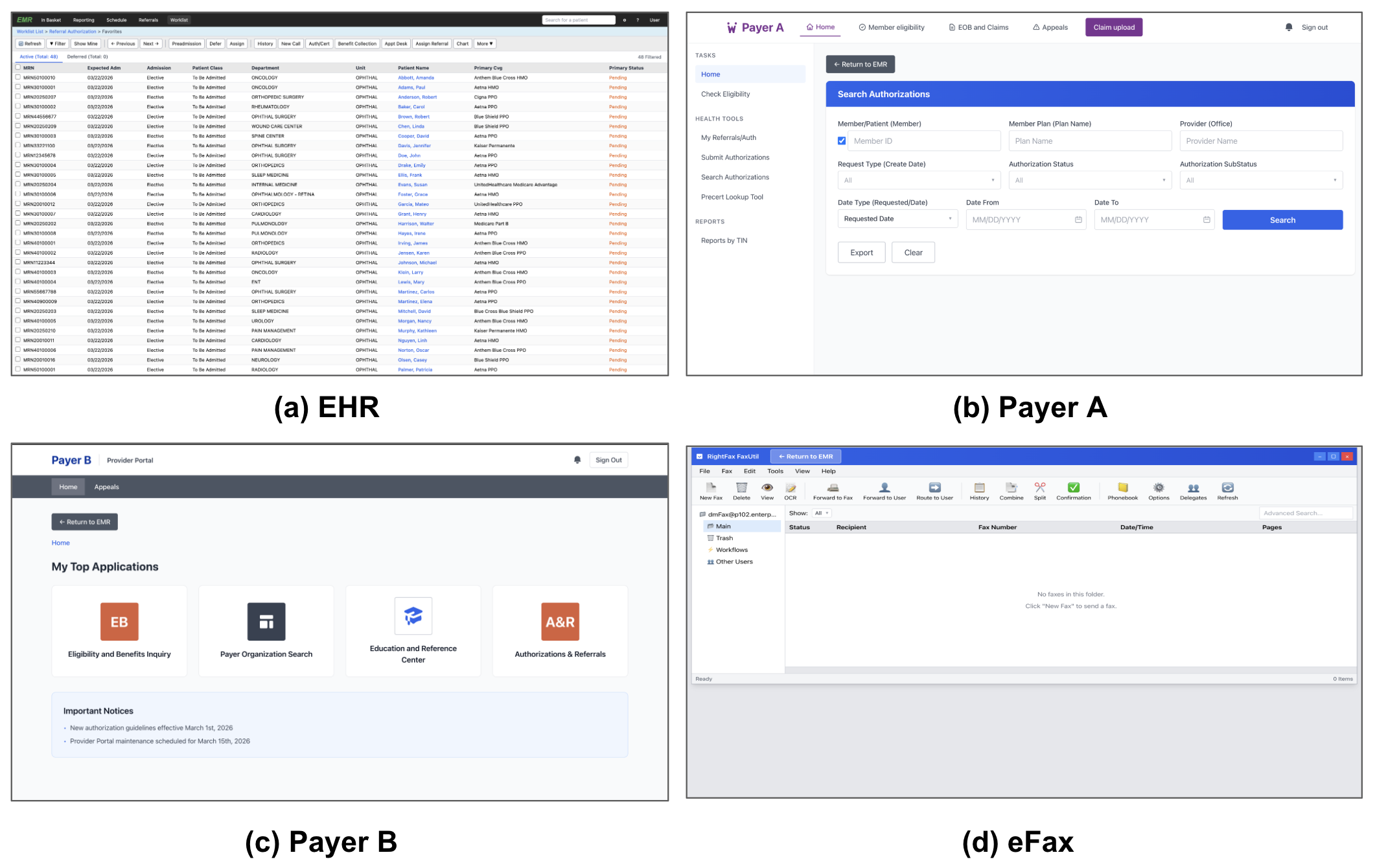}
    \caption{\textsc{HealthAdminBench} contains four environments which mimic commonly utilized applications for administrative healthcare tasks -- (a) an EHR inspired by Epic, (b) a payer portal inspired by Anthem, (c) a portal inspired by Availity, and an eFax inspired by RightFax. These environments are implemented as websites following the REAL framework \cite{garg2025realbenchmarkingautonomousagents}}
    \label{fig:portal_screenshots}
\end{figure}

\vspace{-0.5em}
\subsection{Environments}
\label{subsec:web_envs}
\vspace{-0.5em}

\textsc{HealthAdminBench} includes four deterministic web environments that represent core revenue-cycle systems: an EHR-like viewer with multiple task-specific views (e.g., prior authorization, denials, and DME worklists), two payer authorization portals with distinct layouts and policies, and a fax submission portal. These environments were designed based on over 100 hours of observation of real administrative tasks, during which staff routinely transfer information across multiple systems.

Each environment is implemented as a sandboxed web application with typed inputs, schema validation, policy checks, and realistic failure modes (e.g., missing required fields). Environments are initialized with synthetic patient data to support task execution. Although both the data and environments are synthetic, they are designed to enforce task logic consistent with real payer policies and documentation requirements. We abstract away deployment-specific access constraints such as CAPTCHAs, MFA, and session timeouts in order to isolate workflow execution in a controlled, reproducible setting. Agents must combine chart interpretation with structured form completion under heterogeneous portal constraints.

\vspace{-0.5em}
\subsection{Observation Space}
\vspace{-0.5em}

The observation function $O(s_t)$ defines the information available to the agent at timestep $t$. \textsc{HealthAdminBench} supports observation modalities that reflect realistic enterprise computer use. Specifically, observations may consist of: (i) a rendered screenshot of the current browser viewport, or (ii) a structured interface representation derived from the accessibility tree, in which interactive elements are indexed by stable \texttt{data-testid} identifiers. This representation provides a controlled, reproducible abstraction for element-level interaction. Although such identifiers are typically not exposed in real-world EHR and payer systems, they approximate the structured grounding available through accessibility layers or instrumentation.

The \texttt{data-testid}-indexed representation follows prior computer-use benchmarks \citep{garg2025realbenchmarkingautonomousagents} and allows agents to reference UI elements symbolically without reasoning over pixel coordinates. This aligns with language-based agents that operate over structured or textual inputs. In contrast, screenshot-based observations require visual grounding, where agents must infer both the correct action and its location on the screen.

\vspace{-0.5em}
\subsection{Action Space}
\vspace{-0.5em}

We define a finite action space that captures user-level keyboard and mouse operations. Actions are expressed in natural language at the level of user intent (e.g., ``click submit'', ``enter text''), and are mapped to executable browser operations. For agents with access to the accessibility tree, actions operate on interface elements addressed by their \texttt{data-testid} identifiers, enabling symbolic selection and manipulation of UI components. Agents with screenshot-only observations use the same action set but must additionally infer the spatial location of each action.

The action space includes standard browser interactions such as clicking buttons, entering text, selecting dropdown options, toggling checkboxes, uploading or downloading files, and scrolling. A complete enumeration of supported actions is provided in Table~\ref{tab:action-space}. For Claude Opus 4.6 CUA and GPT-5.4 CUA, however, we use each vendor’s native computer-use tooling rather than this custom action space. These agents operate on the same underlying browser session and environment state as our harness-based agents, and are evaluated identically based on subtask completion in the resulting portal state.

\vspace{-0.5em}
\subsection{Tasks}
\label{subsec:tasks}
\vspace{-0.5em}

\textsc{HealthAdminBench} contains 135 tasks spanning the four environments in Section~\ref{subsec:web_envs}. Following MedHELM’s Administration and Workflow category, we group tasks into three administrative task types, each corresponding to a common revenue-cycle workflow: Prior Authorization, Appeals and Denials Management, and Durable Medical Equipment (DME) Order Processing. Each task is decomposed into fine-grained, verifiable subtasks.

\vspace{-0.5em}
\begin{enumerate}
    \item \textbf{Prior Authorization (60 tasks).} Tasks range from verifying eligibility and documentation to submitting authorization requests via payer portals. More complex tasks require gathering EHR data (e.g., diagnosis codes, NPIs), completing structured forms, and occasionally performing clinical reasoning (e.g., dosage calculations or identifying submission-blocking errors).
    \begin{itemize}
        \item \textit{Example (easy): Determine whether prior authorization is required for a referral and document the outcome.}
        \item \textit{Example (hard): Submit a prior authorization request via a payer portal and record the authorization in the EHR.}
    \end{itemize}

    \item \textbf{Appeals and Denials Management (60 tasks).} Tasks involve reviewing denied claims, determining appropriate resolution (e.g., write-off, resubmission, appeal), and documenting findings. Harder tasks require interacting with payer portals, filtering workqueues, and handling ambiguous or high-value cases.
    \begin{itemize}
        \item \textit{Example (easy): Review a denial and select the correct triage disposition.}
        \item \textit{Example (hard): Identify a high-value denial, file an appeal with supporting documentation, and document the investigation.}
    \end{itemize}

    \item \textbf{DME Order Processing (15 tasks).} Tasks involve retrieving required documentation from the EHR, submitting it to suppliers (e.g., via fax), and recording outcomes. More complex cases introduce constraints such as missing or invalid documentation that should block completion.
    \begin{itemize}
        \item \textit{Example (easy): Download documents, fax to supplier, and document completion.}
        \item \textit{Example (hard): Identify invalid or missing documentation and halt processing.}
    \end{itemize}
\end{enumerate}

These task types collectively capture key failure modes in administrative automation, including incomplete information gathering, incorrect portal interactions, and failure to halt execution under invalid conditions.

\vspace{-1em}
\paragraph{Expert Validation.}
All 135 tasks were designed in collaboration with practicing healthcare administrators. To further validate the task set, we conducted a formal audit on a stratified sample of 40 tasks from Prior Authorization (20) and Appeals and Denials Management (20), spanning all difficulty levels. Each sampled task was independently reviewed by two revenue cycle domain experts with decades of combined experience working at large hospitals, who assessed whether the task goal was realistic, the required steps reflected actual practice, and the success criteria were appropriate. 

Reviewer feedback focused primarily on domain-specific conventions rather than core task design, supporting the validity of the underlying design process. In Prior Authorization, reviewers refined terminology and payer plan types to better reflect real authorization requirements. In Appeals and Denials Management, they identified missing constraints and verification steps (e.g., required routing rules or eligibility checks). These patterns were applied systematically across related tasks beyond those directly reviewed. DME Order Processing tasks follow a more standardized document-submission workflow and were validated through iterative testing rather than formal expert review.

\vspace{-0.5em}
\subsection{Subtasks}
\label{section:subtasks}
\vspace{-0.5em}

Each task in \textsc{HealthAdminBench} is associated with a fixed set of subtasks that define task completion. An agent execution is considered successful only if it satisfies all subtasks, reflecting the high-stakes nature of administrative processes where even minor omissions can invalidate an entire submission.

In total, \textsc{HealthAdminBench} contains 1,698 distinct subtasks. Of these, 1,177 are evaluated using deterministic checks (JMESPath queries over portal state), while 521 require LLM-based evaluation (GPT-5.4), where a judge model assesses whether the agent’s free-text outputs (e.g., triage notes or appeal rationales) satisfy predefined rubrics. We validate the LLM judge on a stratified sample of 60 LLM-judged subtasks from Claude Opus 4.6, finding 93.3\% agreement (Appendix~\ref{sec:llm_judge_evaluation}). All verifiers return strict binary pass/fail outcomes, with no partial credit.

We group subtasks into six subtask types:
\vspace{-0.5em}
\begin{enumerate}
    \item \textbf{Information Retrieval (419 subtasks).} Verifies that the agent navigates to the appropriate interfaces and retrieves required information, such as diagnoses, coverage details, remittance codes, or appeal deadlines.
    \item \textbf{Documentation (515 subtasks).} Verifies that the agent records task-relevant information in clinical notes, including justification, coverage rationale, triage analysis, or disposition summaries.
    \item \textbf{Form Completion (292 subtasks).} Verifies that the agent correctly completes structured forms in payer portals or eFax systems, such as entering member IDs, diagnosis codes, or clinical indications.
    \item \textbf{Task Resolution (200 subtasks).} Verifies that the agent executes terminal actions, such as submitting a form, selecting a triage disposition, sending a fax, or clearing a referral.
    \item \textbf{Document Handling (149 subtasks).} Verifies that the agent correctly transfers documents across environments, such as downloading clinical notes from the EHR and attaching them to a payer portal submission or fax.
    \item \textbf{Clinical Reasoning (123 subtasks).} Verifies that the agent applies domain knowledge to make appropriate clinical or operational decisions, such as identifying diagnosis--procedure mismatches, detecting expired authorizations, or determining when conservative treatment must precede imaging authorization.
\end{enumerate}

These subtask types capture distinct capabilities required for successful execution of administrative tasks, spanning information gathering, structured interaction, and decision-making. Table~\ref{tab:subtask-coverage} summarizes the distribution of subtask types across the three administrative task types, highlighting systematic differences in the capabilities required for each setting (e.g., higher document handling in DME and higher reasoning demands in appeals).

\begin{table}[h]
\centering
\footnotesize
\caption{For each task type, the percentage of tasks that include at least one subtask from each subtask type.}
\label{tab:subtask-coverage}
\setlength{\tabcolsep}{4pt}
\renewcommand{\arraystretch}{0.95}
\begin{tabular}{lrrr}
\toprule
\textbf{Subtask} & \textbf{Prior Authorization} & \textbf{Appeals \& Denials Management} & \textbf{DME} \\
\midrule
Information Retrieval & 93.3\% & 100.0\% & 33.3\% \\
Documentation & 100.0\% & 100.0\% & 100.0\% \\
Form Completion & 58.3\% & 3.3\% & 66.7\% \\
Task Resolution & 98.3\% & 96.7\% & 100.0\% \\
Document Handling & 53.3\% & 23.3\% & 100.0\% \\
Clinical Reasoning & 60.0\% & 71.7\% & 6.7\% \\
\bottomrule
\end{tabular}
\end{table}

\vspace{-0.5em}
\subsection{Memory}
\label{subsec:scratch}
\vspace{-0.5em}

\textsc{HealthAdminBench} includes tasks that require multi-step interaction across multiple interfaces, often exceeding the effective context length of current computer-use agents. As a result, agents cannot condition on the full history of prior actions and observations at each step. To address this, we provide a persistent scratch space that serves as external memory. At each step, the agent may write task-relevant information (e.g., retrieved values, intermediate decisions, or partial progress) to this memory. The stored information is then provided back to the agent in subsequent steps, enabling stateful reasoning over long-horizon tasks.

\vspace{-0.5em}
\subsection{Prompting Strategy}
\label{section:promts}
\vspace{-0.5em}
Agents require prompts that describe the environment, including the available actions, observation modalities, and memory interface. We consider two prompting strategies that vary in the level of domain-specific guidance provided to the agent.
In the \textbf{Task Description} setting, the prompt includes only a description of the environment and a high-level description of the task. In the \textbf{Task Description + Portal Guidance} setting, we augment this prompt with general guidance on performing administrative tasks, including portal navigation patterns and common operational constraints. We additionally use a \textbf{Task-Specific Step-by-Step} prompting for development and trajectory collection. Unlike portal guidance, it explicitly specifies the intended workflow steps and is not used in the primary benchmark evaluation. Full prompt templates and examples are provided in Appendix~\ref{sec:prompting}.

\vspace{-0.5em}
\section{Results}
\vspace{-0.5em}

We evaluate five frontier models on \textsc{HealthAdminBench}: three closed-source models (Claude Opus 4.6, GPT-5.4, and Gemini 3.1 Pro) and two open-source models (Kimi K2.5 and Qwen-3.5-27B). From these, we construct seven agent configurations by pairing models with either our standardized harness (Section~\ref{sec:benchmark}) or native computer-use agent (CUA) implementations, where available. In particular, we evaluate Claude Opus 4.6 and GPT-5.4 using their respective CUA systems—Anthropic’s computer-use tool \cite{anthropic2025computeruse} and OpenAI’s computer-use loop \cite{openai2025computeruse}—to assess the impact of system-level orchestration. Unless otherwise noted, all results use screenshot-only observations with Task Description + Portal Guidance prompting.

\begin{figure}[htbp!]
  \centering
  \includegraphics[width=0.85\linewidth]{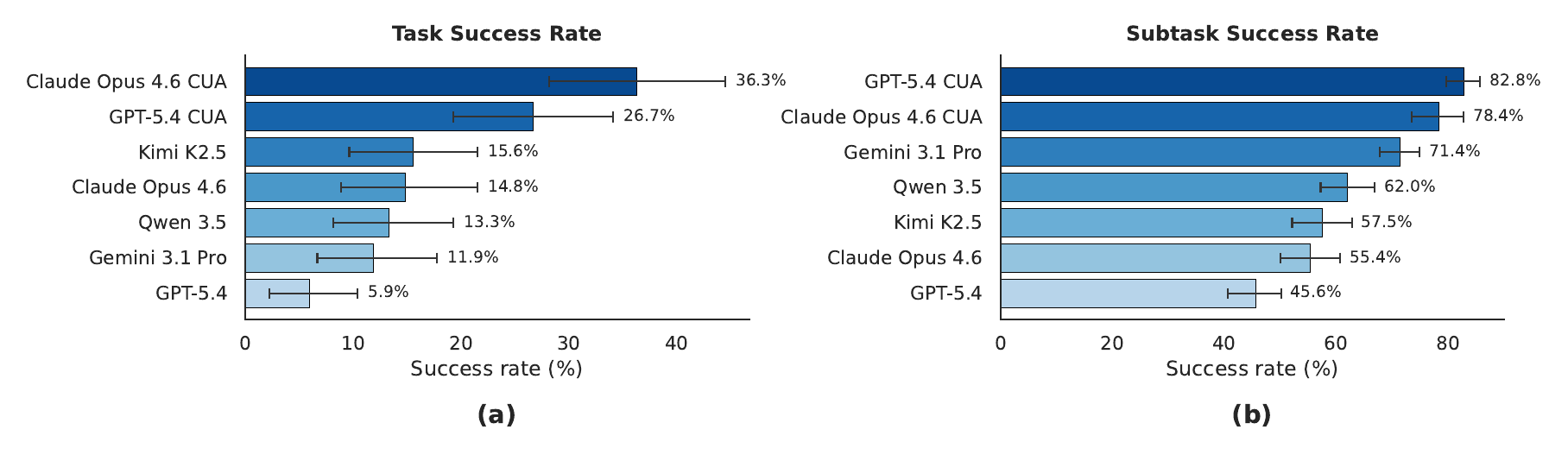}
  \caption{Performance of evaluated agents on \textsc{HealthAdminBench}, reported as (a)  task success rate and (b) subtask success rate. 95\% test-set bootstrap confidence intervals are noted with error bars.}
  \label{fig:leaderboard}
\end{figure}

Figure~\ref{fig:leaderboard} reports overall performance. We track two metrics: \emph{subtask success}, the fraction of subtasks completed correctly, and \emph{task success}, which requires passing \emph{all} subtasks within a task. Since many tasks involve 15 or more subtasks (Section~\ref{section:subtasks}), task success represents a strict end-to-end measure of reliability. We observe a substantial gap between agents using our standardized harness and those using native computer-use systems (CUAs), highlighting the importance of system-level orchestration: Claude Opus 4.6 CUA achieves the highest task success rate, while GPT-5.4 CUA attains the highest subtask success rate. To assess whether the benchmark can reliably distinguish agents, we conduct pairwise head-to-head comparisons (Appendix~\ref{sec:head_to_head}) and find statistically significant differences in 33 of 42 agent pairs, indicating that \textsc{HealthAdminBench} has sufficient scale and granularity for meaningful comparison. Resource usage (steps and API cost) is reported in Appendix~\ref{sec:steps_taken}.

\vspace{-1em}
\subsection{Performance By Task Type}
\vspace{-0.5em}

As described in Section~\ref{subsec:tasks}, tasks are grouped into three administrative task types: Prior Authorization, Appeals and Denials Management, and DME Order Processing. Table~\ref{tab:task_type} shows substantial variation in performance across these task types. Across agents, DME Order Processing achieves the highest success rates, Prior Authorization shows mixed performance, and Appeals and Denials Management remains consistently challenging.

These differences are consistent with variation in task complexity and subtask composition. Compared to DME, Prior Authorization and Appeals and Denials Management involve longer action sequences (Section~\ref{section:subtasks}) and a higher prevalence of Clinical Reasoning and Information Retrieval subtasks (Table~\ref{tab:subtask-coverage}), both of which increase the difficulty of end-to-end execution.

\begin{table}[ht]
    \centering
    \footnotesize
    \setlength{\tabcolsep}{4pt}
    \renewcommand{\arraystretch}{0.95}
    \begin{tabular}{lccc}
        \toprule
        \multirow{2}{*}{\textbf{Agent}} & \multicolumn{3}{c}{\textbf{Task Type}} \\
        \cmidrule{2-4}
        & Prior Authorization & Appeals \& Denials Management & DME \\
        \midrule
        Claude Opus 4.6 CUA & \textbf{38.3\%} & \textbf{25.0\%} & \textbf{73.3\%} \\
        GPT-5.4 CUA & 26.7\% & 18.3\% & 60.0\% \\
        Kimi K2.5 & 18.3\% & 13.3\% & 13.3\% \\
        Claude Opus 4.6 & 16.7\% & 16.7\% & 0.0\% \\
        Qwen 3.5 & 16.7\% & 5.0\% & 33.3\% \\
        Gemini 3.1 Pro & 23.3\% & 3.3\% & 0.0\% \\
        GPT-5.4 & 6.7\% & 3.3\% & 13.3\% \\
        \bottomrule
    \end{tabular}
    \caption{End-to-end task success rates, split by task type.}
    \label{tab:task_type}
\end{table}

\subsection{Performance By Subtask Type}
\vspace{-0.5em}

Subtasks are organized into six subtask types: Information Retrieval, Documentation, Form Completion, Task Resolution, Document Handling, and Clinical Reasoning. Table~\ref{tab:subtask_type} shows substantial variation in performance across these types.

We observe three key trends. First, performance varies markedly by subtask type: Information Retrieval is consistently the easiest, while Clinical Reasoning and Task Resolution are the most challenging. Second, the two CUA agents consistently achieve the strongest performance across subtask types, with GPT-5.4 CUA leading on Documentation, Document Handling, and Clinical Reasoning, and Claude Opus 4.6 CUA leading on Information Retrieval, Form Completion, and Task Resolution. Third, while agent strengths are heterogeneous, non-CUA agents exhibit substantially lower performance, with the largest gaps observed on Documentation, Clinical Reasoning, and Task Resolution.

\begin{table}[ht]
    \centering
    \footnotesize
    \setlength{\tabcolsep}{4pt}
    \renewcommand{\arraystretch}{0.95}
    \begin{tabular}{lcccccc}
        \toprule
        \multirow{2}{*}{\textbf{Agent}} & \multicolumn{6}{c}{\textbf{Subtask}} \\
        \cmidrule{2-7}
        & \shortstack{Information\\Retrieval} & \shortstack{Documentation} & \shortstack{Form\\Completion} & \shortstack{Task\\Resolution} & \shortstack{Document\\Handling} & \shortstack{Clinical\\Reasoning} \\
        \midrule
        Claude Opus 4.6 CUA & \textbf{96.5\%} & 74.3\% & \textbf{88.9\%} & \textbf{70.7\%} & 92.3\% & 61.3\% \\
        GPT-5.4 CUA & 96.1\% & \textbf{84.4\%} & 84.9\% & 66.2\% & \textbf{94.3\%} & \textbf{77.0\%} \\
        Kimi K2.5 & 87.7\% & 45.6\% & 53.2\% & 45.3\% & 58.1\% & 39.3\% \\
        Claude Opus 4.6 & 88.5\% & 43.0\% & 23.4\% & 40.5\% & 49.3\% & 32.0\% \\
        Qwen 3.5 & 87.4\% & 54.0\% & 44.6\% & 51.6\% & 52.0\% & 44.5\% \\
        Gemini 3.1 Pro & 90.2\% & 63.3\% & 87.8\% & 70.3\% & 68.9\% & 55.1\% \\
        GPT-5.4 & 83.0\% & 27.5\% & 36.7\% & 38.5\% & 44.3\% & 22.3\% \\
        \bottomrule
    \end{tabular}
    \caption{Granular success rate for individual subtasks, split by subtask type. Note that one task can include many subtasks.}
    \label{tab:subtask_type}
\end{table}

\vspace{-0.5em}
\subsection{Ablation Experiments: Prompting and Observation}
\label{sec:ablations}
\vspace{-0.5em}

In our primary experiments, we evaluate agents using Task Description + Portal Guidance prompting with screenshot-only observations. To test sensitivity to these design choices, we vary both prompting strategy and observation modality. Figure~\ref{fig:ablation} shows two patterns. First, portal guidance consistently improves task success over task description alone, with larger gains for end-to-end performance. Second, accessibility-tree observations generally outperform screenshot-only observations, especially when combined with portal guidance. Together, these results show that domain-specific prompting and structured UI representations improve agent reliability, although the latter is often unavailable in real deployments. The same trends hold for subtask success, though values are higher across all conditions, reflecting the gap between completing individual steps and completing full workflows. We report the full subtask breakdown in Appendix~\ref{sec:head_to_head} (Figure~\ref{fig:ablation-subtask}).

\vspace{-0.5em}
\subsection{Fine-Tuning on Domain-Specific Data}
\vspace{-0.5em}

We assess domain adaptation by randomly splitting \textsc{HealthAdminBench} tasks into a small training set (100 tasks) and a held-out test set (35 tasks). All domain adaptation experiments are performed with accessibility-tree observations and Task Description only prompting to reduce cost. We generate training trajectories using Task-Specific Step-by-Step prompting and use them as supervision to fine-tune Qwen 3.5, yielding Qwen-3.5-Kinetic-SFT. Evaluated on the held-out test set, this fine-tuned model achieves a task success rate of 40\%, a gain over its base version of +23\%, and outperforms the best-performing frontier model (Claude Opus 4.6) by over +14\%. These preliminary results suggest that even limited domain-specific supervision can substantially improve performance on complex administrative workflows. Full details are provided in Appendix~\ref{sec:finetune}.

\vspace{-0.5em}
\subsection{Qualitative Analysis of Failure Modes}
\vspace{-0.5em}

To understand where agents fail, we manually analyzed trajectories from Claude Opus 4.6 and observed similar patterns across models. Three recurring failure modes account for the majority of errors.

\vspace{-8pt}
\paragraph{Hidden long-term dependencies.}
Many tasks require first gathering information in the EHR and later using it in payer portals or fax systems, often with requirements that are not apparent upfront. While humans can recover by revisiting earlier steps, agents rarely do so and frequently exhaust their step budget, leading to systematic failures.

\vspace{-8pt}
\paragraph{Avoidance of file operations.}
Despite support for file downloads and uploads, agents often avoid or ignore these actions. Since document transfer is essential for prior authorization and DME workflows, this behavior leads to widespread failure in document-handling subtasks.

\vspace{-8pt}
\paragraph{Information loss over long horizons.}
Agents operate under limited context and typically observe only recent steps, causing earlier information to be forgotten unless explicitly stored. Although we provide a scratch space, agents often fail to record key details, leading to downstream errors when later steps depend on information gathered earlier in the workflow.

\begin{figure}[htbp!]
  \centering
  \includegraphics[width=0.9\linewidth]{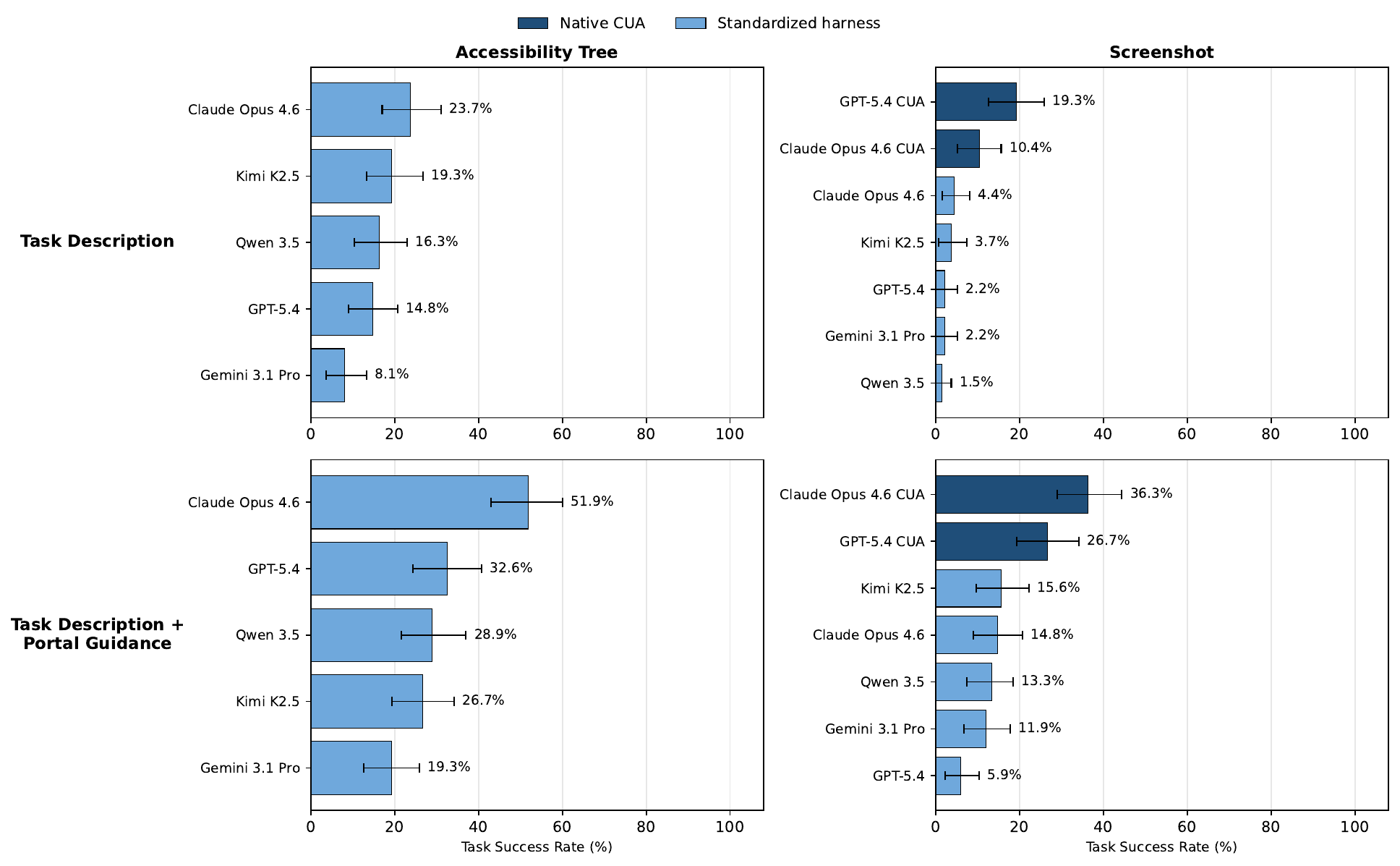}
  \caption{Performance of evaluated agents on \textsc{HealthAdminBench} across prompting and observation settings, reported as task success rate.}
  \label{fig:ablation}
\end{figure}
\vspace{-2em}
\section{Discussion}
\label{sec:discussion}
\vspace{-1em}

\textsc{HealthAdminBench} addresses a key gap in the evaluation of healthcare AI agents by moving beyond static, text-only assessments to realistic administrative workflows that require long-horizon, cross-system interaction. Across all evaluated agents, end-to-end task success remains low—the strongest agent (Claude Opus 4.6 CUA) reaches only 36.3\%—despite much higher subtask success rates. This gap suggests that current computer-use agents can often execute individual actions correctly, yet still struggle to reliably complete full administrative workflows, particularly when tasks require document handling, cross-portal coordination, and sustained state tracking over many steps. At the same time, our fine-tuning results suggest that these failures are not solely a consequence of model capability, but also reflect the scarcity of domain-specific workflow data and training environments.

\vspace{-8pt}
\paragraph{Limitations.}
While our environments and task set capture key sources of structural difficulty in healthcare administrative workflows, they remain simplified proxies for production systems. In particular, the benchmark is static and retrospective: it does not capture longitudinal interface drift, evolving payer policies, or the full heterogeneity of real hospital deployments. Although the accessibility-tree setting enables controlled and reproducible evaluation, real-world portals do not typically expose such clean, stable identifiers, and frequent UI updates may reduce the robustness of agents that rely on them. Agents also operate under limited context windows, we evaluate single-shot trajectories without human-like recovery behavior, and we abstract away deployment constraints such as MFA, CAPTCHAs, and session timeouts. More broadly, while administrative automation offers the potential to reduce cost and burden, premature deployment of agentic systems risks operational errors with financial or clinical consequences, underscoring the need for rigorous evaluation and human oversight.

\vspace{-1em}
\section{Conclusion}
\vspace{-1em}

We introduce \textsc{HealthAdminBench}, the first benchmark for evaluating computer-use agents on realistic healthcare administrative workflows. Our results show that, despite promising subtask-level performance, current frontier agents remain far from reliable on end-to-end tasks, revealing a substantial gap between interface-level competence and deployment-ready workflow automation. By releasing reproducible environments, expert-grounded tasks, and fine-grained evaluation protocols, we aim to provide a foundation for measuring progress toward safer and more effective administrative AI systems in healthcare.

\vspace{-1em}
\section{Acknowledgements}
\vspace{-1em}

We want to thank Kinetic Systems for allowing their employees (Ethan Steinberg, Michael Wornow, Haroun Ahmed, Peter Sterling, and Bravim Purohit) to volunteer their time with Stanford University to contribute to this paper.

\bibliography{iclr2026_conference}
\bibliographystyle{iclr2026_conference}

\newpage
\appendix
\part{Appendix}

\localtableofcontents
\newpage
\section*{Appendix}\label{app:appendix}
\section{Prompting Strategy Details}
\label{sec:prompting}

Reliable prompting is necessary for obtaining stable behavior across long-horizon administrative workflows. Here, we present the prompting design used in \textsc{HealthAdminBench}.

\subsection{Base System Prompts}
\label{subsec:system_prompt}

\begin{table}[h]
\centering
\small
\begin{tabular}{l l | p{8.2cm}}
\toprule
\textbf{Context} & \textbf{Action} & \textbf{Description} \\
\midrule
\multirow{13}{*}{Screenshot Only}
& \texttt{click\_coord(x, y)} & Left-click at pixel coordinates $(x, y)$ (origin at top-left).\\
& \texttt{double\_click\_coord(x, y)} & Double-click at pixel coordinates $(x, y)$. \\
& \texttt{triple\_click\_coord(x, y)} & Triple-click at pixel coordinates $(x, y)$. \\
& \texttt{right\_click\_coord(x, y)} & Right-click at pixel coordinates $(x, y)$. \\
& \texttt{move\_coord(x, y)} & Move the cursor to pixel coordinates $(x, y)$. \\
& \texttt{type\_text(text)} & Type text at the current cursor focus. \\
& \texttt{type\_text\_coord(text, x, y)} & Click at $(x, y)$ then type text at that location. \\
& \texttt{key\_press(key)} & Press a keyboard key or key combination. \\
& \texttt{scroll(dx, dy)} & Scroll by pixel offsets (positive $dy$ scrolls downward). \\
& \texttt{scroll(x, y, dx, dy)} & Move to $(x,y)$ and scroll by pixel offsets. \\
& \texttt{wait(t)} & Pause execution for $t$ seconds. \\
& \texttt{done()} & Signal that the task has been completed. \\
\midrule
\multirow{10}{*}{Accessibility Tree}
& \texttt{click(id)} & Click the UI element referenced by \texttt{data-testid} \texttt{id}. \\
& \texttt{fill(id, text)} & Enter text into an input field. \\
& \texttt{select(id, option)} & Select an option from a dropdown menu by visible label. \\
& \texttt{press(id, key)} & Press a keyboard key or key combination on a specific element. \\
& \texttt{scroll(dir)} & Scroll the page vertically (\texttt{up} or \texttt{down}). \\
& \texttt{back()} & Navigate to the previous page in browser history. \\
& \texttt{download(id)} & Click a download link or button and save the resulting file. \\
& \texttt{upload(id, file)} & Upload a file to a file input element. \\
& \texttt{done()} & Signal that the task has been completed. \\
\bottomrule
\end{tabular}
\caption{Complete action space for HealthAdminBench. Agents with access to the accessibility tree operate on symbolic UI elements referenced by \texttt{data-testid} identifiers, while screenshot-only agents interact with the environment via coordinate-based mouse and keyboard actions.}
\label{tab:action-space}
\end{table}

We use a fixed system prompt to (i) specify the agent role, (ii) define the admissible action space and required syntax (Table~\ref{tab:action-space}), and (iii) enforce a structured output format that includes both the next action and a concise summary of newly observed, task-relevant information. This format provides an explicit task trace that can be carried forward across timesteps (Section~\ref{subsec:scratch}). We additionally include constraints to ensure actions are executable (e.g., only referencing element identifiers that appear in the accessibility tree) and to standardize termination (the agent must emit \texttt{done()} when it believes the objective is complete).

All harness-based agents are queried through standard chat or completion endpoints, without vendor-specific reasoning-mode parameters. The \texttt{THINKING} field is a structured output instruction for an explicit action rationale and task trace, not a special reasoning API. For CUA configurations, the vendor's native computer-use SDK manages the interaction loop internally.

The full system prompt used for screenshot-only and accessibility-tree settings is provided below.

\begin{tcolorbox}[
  breakable,
  colback=gray!5,
  colframe=black!50,
  boxrule=0.5pt,
  arc=2pt,
  left=4pt,
  right=4pt,
  top=4pt,
  bottom=4pt,
  fontupper=\ttfamily\small,
  title={Base System Prompt (Screenshot-Only Setting)}
]
You are an autonomous web agent that can interact with websites by performing actions. \\

Your task is to complete the given objective by analyzing the current page and selecting the appropriate action. \\

AVAILABLE ACTIONS (SCREEN COORDINATES): \\
- click\_coord(x, y) - Left click at pixel coordinates (origin top-left) \\
- double\_click\_coord(x, y) - Double click at pixel coordinates \\
- triple\_click\_coord(x, y) - Triple click at pixel coordinates \\
- right\_click\_coord(x, y) - Right click at pixel coordinates \\
- move\_coord(x, y) - Move mouse to pixel coordinates \\
- type\_text("text") - Type text at the current cursor focus \\
- type\_text\_coord("text", x, y) - Click at (x, y) then type text at that location \\
- key\_press("Enter") - Press a key or key combo (e.g., "Enter", "Ctrl+L") \\
- scroll(dx, dy) - Scroll by pixel offsets (positive dy = down) \\
- scroll(x, y, dx, dy) - Move to (x,y) then scroll by offsets \\
- wait(seconds) - Pause before the next action \\
- done() - Signal that you have completed the objective \\

ACTION FORMAT: \\
You MUST respond with an action AND key information from the current page using this format: \\
THINKING: <think through your past actions, key observations gathered so far, the objective, and the current page to determine the next single action to take to achieve the objective> \\
ACTION: action\_string \\
KEY\_INFO: concise but complete summary of all NEW information from this page potentially relevant to completing the task. Do NOT repeat facts already listed in KEY INFORMATION GATHERED SO FAR unless they have changed. Include specific values (IDs, dates, names, amounts, statuses, codes, credentials). Use a single line with separators like "; " or " | " to retain multiple facts. \\

Examples: \\
- ACTION: click\_coord(420, 318) \\
  KEY\_INFO: Clicking the login button. \\

- ACTION: type\_text("user@example.com") \\
  KEY\_INFO: Typed the text "user@example.com". \\

- ACTION: type\_text\_coord("user@example.com", 420, 318) \\
  KEY\_INFO: Typed the text "user@example.com" after clicking on the input field. \\

- ACTION: scroll(0, 500) \\
  KEY\_INFO: Scrolling to reveal more options. \\

- ACTION: scroll(640, 360, 0, 500) \\
  KEY\_INFO: Moving to center and scrolling down. \\

- ACTION: done() \\
  KEY\_INFO: Task completed. \\

IMPORTANT GUIDELINES: \\
1. Coordinates are in pixels relative to the screenshot \\
2. Use the screenshot to locate UI elements visually \\
3. Prefer clicking UI elements instead of typing URLs \\
4. Complete the objective step by step
\end{tcolorbox}

\begin{tcolorbox}[
  breakable,
  colback=gray!5,
  colframe=black!50,
  boxrule=0.5pt,
  arc=2pt,
  left=4pt,
  right=4pt,
  top=4pt,
  bottom=4pt,
  fontupper=\ttfamily\small,
  title={Base System Prompt (Accessibility Tree Setting)}
]
You are an autonomous web agent that can interact with websites by performing actions. \\

Your task is to complete the given objective by analyzing the current page and selecting the appropriate action. \\

AVAILABLE ACTIONS: \\
- click([id]) - Click an element with the specified identifier (USE THIS FOR NAVIGATION) \\
- fill([id], "text") - Type text into an input field \\
- select([id], "option") - Select dropdown option by visible label (for <select> elements) \\
- press([id], "key") - Press a key or key combination on a specific element \\
- scroll(down) or scroll(up) - Scroll the page \\
- back() - Go back to the previous page (browser back button) - USE THIS to return from external portals \\
- download([id]) - Click a download button/link and save the file (use for downloading documents like auth letters) \\
- upload([id], "filename") - Upload a previously downloaded file to a file input (use "last" for the most recent download) \\
- done() - Signal that you have completed the objective \\

ACTION FORMAT: \\
You MUST respond with an action AND key information from the current page using this format: \\
THINKING: <think through your past actions, key observations gathered so far, the objective, and the current page to determine the next single action to take to achieve the objective> \\
ACTION: action\_string \\
KEY\_INFO: concise but complete summary of all NEW information from this page potentially relevant to completing the task. Do NOT repeat facts already listed in KEY INFORMATION GATHERED SO FAR unless they have changed. Include specific values (IDs, dates, names, amounts, statuses, codes, credentials). Use a single line with separators like "; " or " | " to retain multiple facts. \\

Examples: \\
- ACTION: click([login-button]) \\
  KEY\_INFO: Found login form with username and password fields. \\

- ACTION: fill([email-input], "user@example.com") \\
  KEY\_INFO: None - just filling the form. \\

- ACTION: download([download-auth-letter]) \\
  KEY\_INFO: Downloading auth letter to attach as supporting documentation. \\

- ACTION: upload([file-upload-input], "last") \\
  KEY\_INFO: Uploading the previously downloaded auth letter. \\

- ACTION: done() \\
  KEY\_INFO: Task completed - note added and referral cleared. \\

IMPORTANT GUIDELINES: \\
1. Always extract element identifiers from the PAGE ELEMENTS section \\
2. Only use identifiers that are explicitly shown in PAGE ELEMENTS (e.g., [id]) \\
3. Do not invent or guess identifiers \\
4. In axtree\_only mode, PAGE ELEMENTS already includes the full page; scrolling rarely reveals new elements \\
5. If an element is not in PAGE ELEMENTS, try checking other tabs or sections \\
6. Complete the objective step by step \\
7. Call done() only when the entire objective is accomplished
\end{tcolorbox}

\subsection{Task Description Prompt Amendments}
\label{subsec:zero_shot_prompt}

To evaluate agent performance under minimal guidance, we adopt a prompting setting in which the agent receives only the base system prompt and the task goal. This setting is designed to measure the extent to which agents can infer appropriate strategies for administrative workflows from the environment alone.

\subsection{Portal Guidance Prompt Amendments}
\label{subsec:informative_user_prompt}

As healthcare administrative portals and the associated multi-portal workflows are nontrivial, we additionally evaluate agents under a more informative prompting setting. In this regime, the system prompt is augmented with domain-specific procedural guidance for interacting with each portal, including how to extract credentials, download and attach documents, navigate patient records, and manage cross-portal workflows. Importantly, these prompts provide general operational guidance rather than step-by-step task solutions, preserving the need for agents to reason about task execution. \\

Portal guidance prompt amendments are assembled per-task from modular, workflow-specific hint blocks. Each task receives: (1) action syntax notes specific to the observation mode (accessibility tree or coordinate), (2) shared execution rules covering document transfer and multi-portal navigation, (3) EMR-specific guidance for the relevant workflow module (prior authorization worklist, denials workqueue, or DME orders), and (4) payer portal or fax portal guidance if the task involves cross-portal navigation. \\

Below, we present a representative portal guidance prompt amendment for a prior authorization task using the accessibility tree setting. The complete set of prompt modules for all workflows and portals is available in the codebase. 

\begin{tcolorbox}[
  breakable,
  colback=gray!5,
  colframe=black!50,
  boxrule=0.5pt,
  arc=2pt,
  left=4pt,
  right=4pt,
  top=4pt,
  bottom=4pt,
  fontupper=\ttfamily\small,
  title={Portal Guidance System Prompt Amendment}
]

\textbf{ACTION SYNTAX (DOM / testid mode):} \\
- Click: click([testid]) $|$ Fill: fill([testid], "value") $|$ Back: back() \\
- Dropdowns are custom (NOT native \texttt{<select>}). Two steps required: \\
\hspace*{1em}1. click([dropdown-testid]) to open the options list. \\
\hspace*{1em}2. click([dropdown-testid-option-\{value\}]) to select the desired option. \\
\hspace*{1em}Do NOT use select() --- it will not work. \\
- Dates: fill the text field with MM/DD/YYYY. \\

\textbf{DOCUMENT TRANSFER:} \\
- Download all required documents in EMR BEFORE navigating to any payer or fax portal. \\
- To open a document: click the ``View $\rightarrow$'' button on the RIGHT side of the document row. Do NOT click the document name/title. \\
- On the viewer page, click Download. Then click ``$<$ Back'' EXACTLY ONCE to return to the referral page. \\
- Downloaded documents automatically appear in the ``Available Documents from EMR'' section inside payer portals. \\

\textbf{MULTI-PORTAL WORKFLOW:} \\
- Gather all information and download all documents in EMR before leaving for a payer portal. \\
- Navigate to the payer portal via the Coverages tab portal link. \\
- Never navigate back mid-form --- progress will be lost. \\
- After portal submission, use ``Return to EMR'' to navigate back, then add a note with the confirmation number. \\

\textbf{EMR WORKLIST:} \\
- Find the referral by scrolling the patient list or using the search field. Click the patient/referral row to open it. \\
- Tab navigation --- all evaluated, must visit in this order: \\
\hspace*{1em}1. [EVALUATED] Diagnoses tab $\rightarrow$ record all ICD-10 codes. \\
\hspace*{1em}2. [EVALUATED] Services tab $\rightarrow$ record all CPT/HCPCS codes. \\
\hspace*{1em}3. Referral tab $\rightarrow$ capture referral details. \\
\hspace*{1em}4. [EVALUATED] General tab $\rightarrow$ scroll to Documents section $\rightarrow$ for each required doc, click ``View $\rightarrow$'' then Download. \\
\hspace*{1em}5. [EVALUATED] Coverages tab $\rightarrow$ capture payer credentials and portal link. SCROLL DOWN to find the ``Open Portal'' button. \\
- After returning from a payer portal: scroll to Communications $\rightarrow$ Add Note $\rightarrow$ fill subject and content (include the confirmation number) $\rightarrow$ Save $\rightarrow$ Clear from Worklist. \\

\textbf{PAYER A:} \\
- Eligibility check: Click ``Member Eligibility'' tab $\rightarrow$ fill Member ID, First Name, Last Name, DOB $\rightarrow$ submit. \\
- Submit prior authorization: Provider lookup $\rightarrow$ Request Type $\rightarrow$ Patient lookup $\rightarrow$ Diagnoses (add each ICD-10) $\rightarrow$ CPT codes (add each) $\rightarrow$ Clinical indication $\rightarrow$ Attach docs $\rightarrow$ Submit $\rightarrow$ capture confirmation ID $\rightarrow$ return to EMR. \\
- Look up / dispute a claim: Appeals tab $\rightarrow$ enter member/claim ID $\rightarrow$ Search $\rightarrow$ click claim $\rightarrow$ Dispute Claim $\rightarrow$ fill rationale $\rightarrow$ attach docs $\rightarrow$ Submit. \\
\end{tcolorbox}

Equivalent modules exist for Payer B, the fax portal, the denials workqueue, and DME orders; see the codebase for the complete set.

\subsection{Task-Specific Prompt Amendments}

To assess whether each task in \textsc{HealthAdminBench} is, in principle, completable by a computer-use agent, we introduce \textit{task-specific prompt amendments} as the most detailed level of system prompting. This setting extends the informative prompting instructions described in Section~\ref{subsec:informative_user_prompt} by additionally providing explicit, step-by-step natural-language instructions for completing a particular task.

Since such detailed procedural guidance is unlikely to be available in realistic deployments of agentic systems, we use task-specific prompt amendments solely for validation and debugging. Results obtained under this setting are therefore \textbf{not} included in the benchmark's reported performance metrics. Instead, this regime serves to verify task correctness, environment consistency, and the feasibility of successful task completion.

Below, we present an example of a task-specific prompt amendment for a DME order processing task that requires the agent to identify an expired document and halt processing.

\begin{tcolorbox}[
  breakable,
  colback=gray!5,
  colframe=black!50,
  boxrule=0.5pt,
  arc=2pt,
  left=4pt,
  right=4pt,
  top=4pt,
  bottom=4pt,
  fontupper=\ttfamily\small,
  title={Task-Specific System Prompt Amendment (DME Order Processing)}
]
\textbf{MANDATORY STEP-BY-STEP GUIDE --- FOLLOW EVERY STEP IN ORDER:} \\
These steps are not suggestions. Each step must be executed. \\
Steps marked [REQUIRED] or [EVALUATED] are directly scored by the evaluation system. Skipping them results in 0 points for that criterion, even if the correct outcome is achieved. \\
Do NOT skip any step, even if the information appears already visible. \\

1. On the DME Orders page (\texttt{/emr/dme}), click on patient Garcia, Sofia \\
2. The referral opens with the Orders tab (Active sub-tab by default). Review the order details and note the DME supplier (Option Care Health). \\
3. Click the \texttt{Chart Review} tab in the top navigation bar \\
4. In Chart Review, open the Face-to-Face Evaluation document --- observe that it is dated April 2025, more than 6 months before the order date \\
5. This indicates the F2F evaluation is expired per Medi-Cal policy (must be within 6 months) \\
6. Do NOT open the fax portal or send any fax \\
7. Do NOT clear the referral from the worklist \\
8. Click the \texttt{Notes} tab in the top navigation bar \\
9. In the Notes tab (right panel, Edit Note), enter a subject and a detailed progress note documenting: patient Sofia Garcia; enteral feeding pump order; expired F2F evaluation (April 2025); Medi-Cal requirement that F2F be within 6 months; need for updated F2F from Dr.\ Emily Foster; order on hold; no fax sent to Option Care Health \\
10. Click \texttt{Sign} to save the note
\end{tcolorbox}

\subsection{User Prompts}

The user prompt presents the agent with the current environment state, including the task objective, interaction metadata, and the most recent observation. To ensure reproducibility regardless of when the benchmark is run, all tasks use a fixed benchmark date (February 25, 2026). This date is injected into every user prompt so that agents reason about deadlines, appointment dates, and document expiration consistently across evaluations.

Specifically, the prompt contains the task description, the fixed benchmark date, step count, current URL, recent action--observation history, and the current observation in the form of the accessibility tree and/or a browser screenshot, depending on the observation mode. At the end of each prompt, the agent is explicitly instructed to produce a thinking trace, the next action, and a summary of any newly observed, task-relevant information. The harness also includes loop detection: if the agent repeats the same action multiple times, a warning or critical alert is injected into the prompt to encourage the agent to try a different strategy.

Below, we show the scaffolding of the user prompt for the setting in which both the accessibility tree and screenshots are provided.

\begin{tcolorbox}[
  breakable,
  colback=gray!5,
  colframe=black!50,
  boxrule=0.5pt,
  arc=2pt,
  left=4pt,
  right=4pt,
  top=4pt,
  bottom=4pt,
  fontupper=\ttfamily\small,
  title={User Prompt}
]
OBJECTIVE: <task description> \\
Treat the current benchmark date as Wednesday, February 25, 2026. \\

CURRENT URL: <current browser URL> \\
STEP: <step count> \\

[Screenshot of current page is attached] \\

RECENT ACTIONS AND KEY OBSERVATIONS (most recent last): \\
\hspace*{1em}ACTION: <previous action 1> \\
\hspace*{1em}| OBSERVATION: <key info from that step> \\
\hspace*{1em}ACTION: <previous action 2> \\
\hspace*{1em}| OBSERVATION: <key info from that step> \\

PAGE ELEMENTS (use identifiers shown in [brackets]): \\
<accessibility tree content, if included in the observation mode> \\

Analyze the current page and objective. What is the next single action to take? \\
Respond with: \\
THINKING: <think through your past actions, key observations gathered so far, the objective, and the current page to determine the next single action to take to achieve the objective> \\
ACTION: <your action> \\
KEY\_INFO: <concise but complete summary of all NEW information from this page potentially relevant to completing the task; do NOT repeat facts already listed above unless they have changed; include IDs, dates, names, amounts, statuses, codes, credentials; use a single line with separators like ``;'' or ``|''> \\
\end{tcolorbox}
\section{LLM Judge Evaluation}
\label{sec:llm_judge_evaluation}

Our benchmark relies heavily on an LLM judge (GPT 5.4), with 521 (30.7\%) subtasks evaluated by them. As such, it's important to understand the accuracy of our LLM judge, which we measure by comparing the judge against humans. We compare four human reviewers to our chosen LLM judge, GPT 5.4. Our judge evaluation dataset consists of 60 randomly selected LLM judge subtask submissions from Claude Opus 4.6 (with Task Description + Portal Guidance prompting and axtree observations). We perform stratified sampling, selecting 30 subtasks where the LLM judge claimed that Claude Opus 4.6 failed and 30 subtasks where the LLM judge claimed that Claude Opus 4.6 succeeded. Each of the 60 subtask submissions is reviewed by 2 human reviewers for a total of 120 human reviews (distributed equally among the 4 human reviewers). Human reviewers evaluated whether Claude Opus 4.6 succeeded or failed the subtasks without reference to the LLM judge's result or reasoning.

The human reviewers unanimously agree that Claude Opus 4.6 succeeded on 30 subtasks and failed on 26, but were conflicted about the remaining 4. This corresponds to an agreement rate of 93.3\% between the human reviewers. The 95\% confidence interval for the human-human agreement rate is 83.8\% to 98.15\%. Each of the 120 human reviews can also be compared to the LLM judge. Humans agree with the LLM judge on 112 subtasks and disagree on the remaining 8. This corresponds to an agreement rate of 93.3\% between the human reviewers and the LLM judge. The 95\% confidence interval for the human-judge agreement rate is 87.4\% to 97.1\%.

These high agreement rates strongly indicate that our rubrics are reliable (i.e., that different independent people can reach the same conclusion) and that GPT 5.4 serves as an adequate LLM judge.
\section{Resources Used: Steps Taken and Cost}
\label{sec:steps_taken}

\begin{figure}[H]
    \centering
    \includegraphics[width=0.9\linewidth]{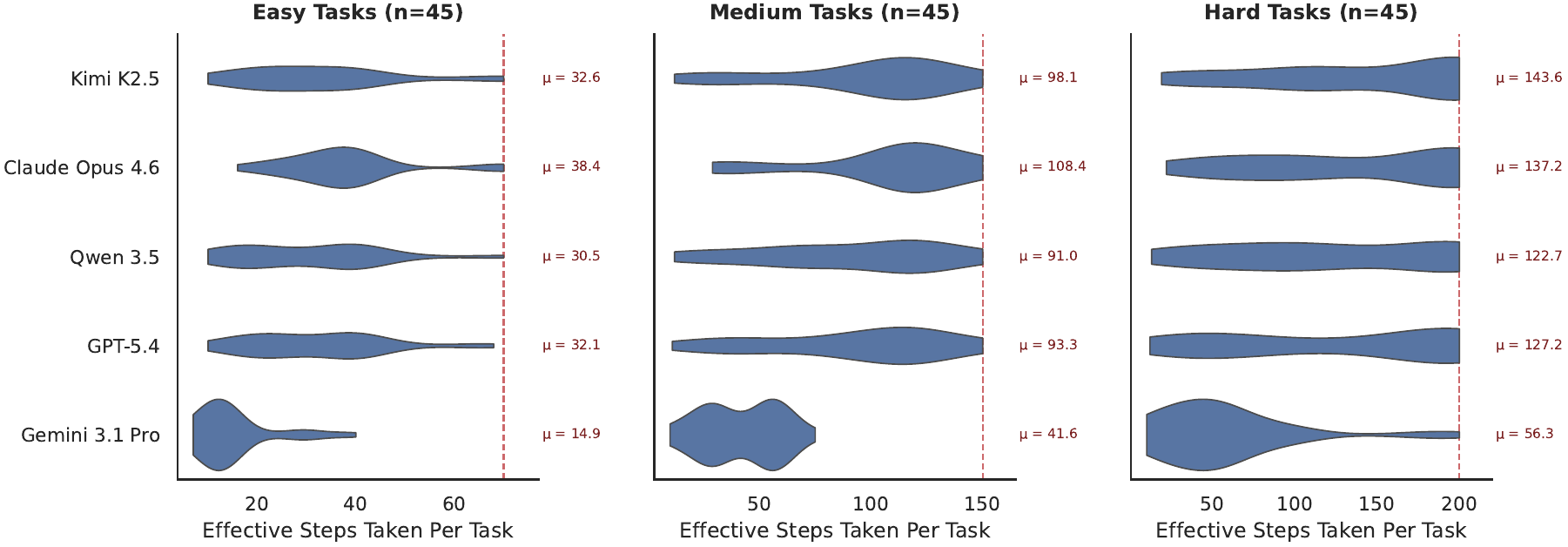}
    \caption{Violin plots indicating the distribution of steps taken by each agent for easy, medium, and hard tasks. The average number of steps for each agent is printed to the right of each violin. A dotted red line indicates the maximum steps each model was allowed to take before being terminated.}
    \label{fig:steps_taken}
\end{figure}

Resource usage is an important consideration for computer-use agents, as each interaction incurs both latency and API cost. Figure \ref{fig:steps_taken} shows the number of steps taken by each agent. We find that Gemini 3.1 Pro uses the fewest steps. Qualitative analysis of Gemini 3.1 Pro traces indicates that this is due to Gemini 3.1 Pro's tendency to give up on tasks when it struggles, whereas other models keep trying until they run out of steps.  GPT-5.4 uses fewer steps because it often performs many steps per turn (for example, click then type actions). For resource accounting, we count a step as a single turn in which the model is provided with input and asked to take actions.

\begin{figure}[H]
    \centering
    \includegraphics[width=0.9\linewidth]{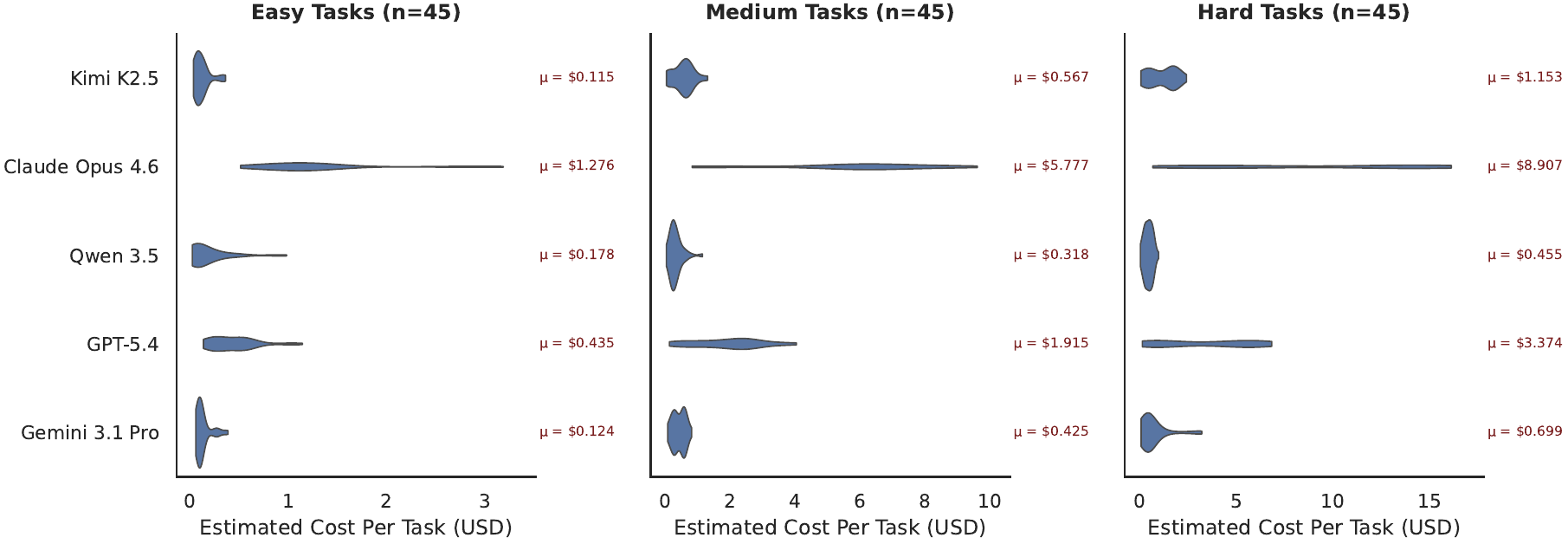}
    \caption{Violin plots indicating the cost taken by each agent for easy, medium, and hard tasks on HealthAdminBench. The average cost is printed to the right of each violin. CUA agents have not been plotted due to harness limitations.}
    \label{fig:cost_used}
\end{figure}

We can also compare agents by looking at the agent API costs used while performing tasks. Figure \ref{fig:cost_used} shows the agent API costs.  Note that Anthropic CUA and OpenAI CUA are not plotted due to inconsistent cost accounting in their respective harness code, which makes a fair comparison difficult. We find that most agents are comparable in cost, except for Claude Opus 4.6, which is very expensive to run.

\section{Additional Performance Breakdowns}
\label{sec:head_to_head}

\subsection{Subtask Success Rate by Prompting \& Observation}
\begin{figure}[h]
    \centering
    \includegraphics[width=\textwidth]{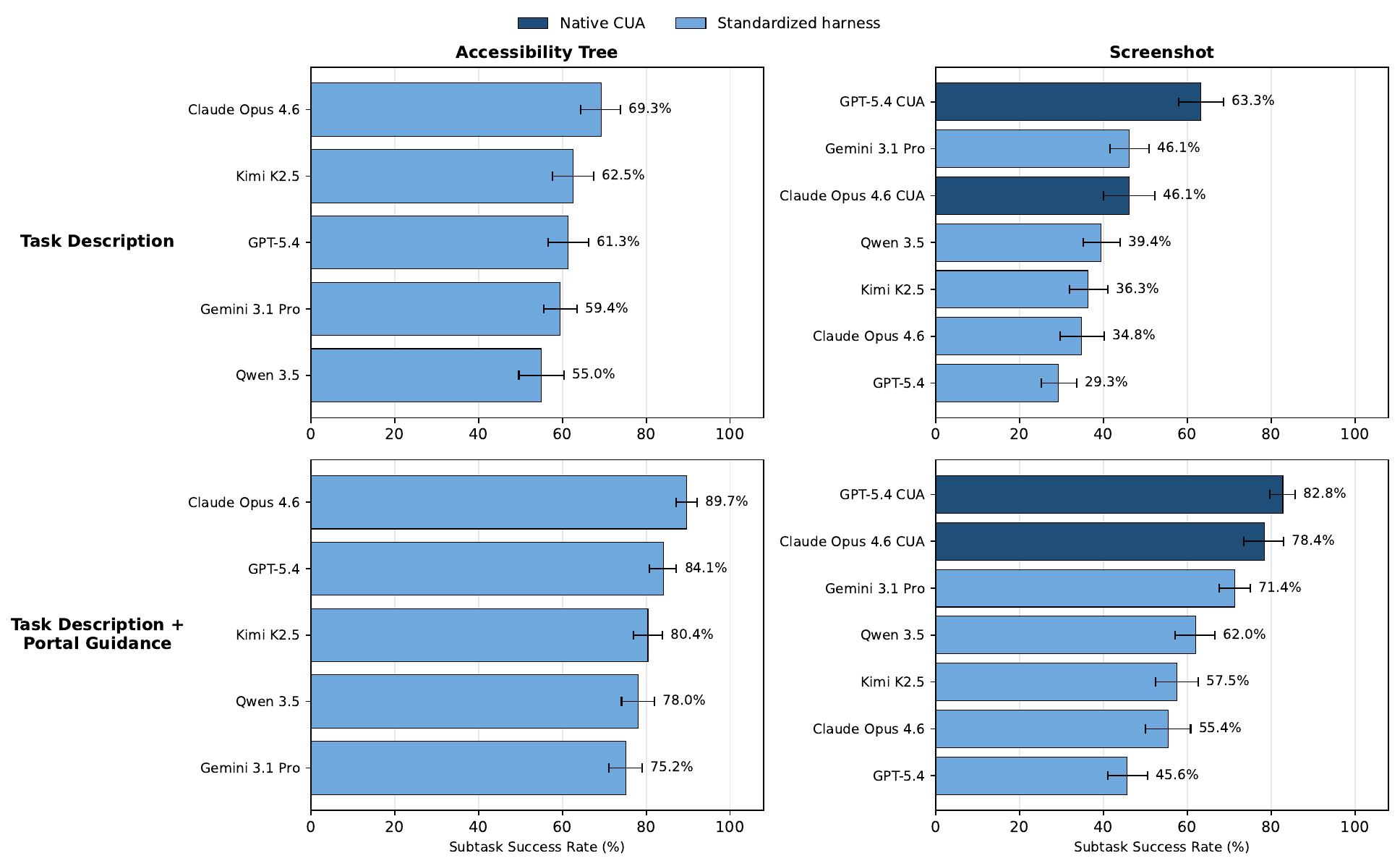}
    \caption{Subtask success rate of evaluated agents on \textsc{HealthAdminBench} across prompting strategy (rows) and observation modality (columns). Error bars show 95\% bootstrap confidence intervals on the test set. Native CUA agents (dark bars) are evaluated only in screenshot mode. This figure complements Figure~\ref{fig:ablation}, which reports the same experiments by task success rate.}
    \label{fig:ablation-subtask}
\end{figure}

Figure~\ref{fig:ablation-subtask} reports subtask success rate across the same prompting $\times$ observation grid as Figure~\ref{fig:ablation}. The same ordering holds: Portal Guidance improves over Task Description alone, and Accessibility Tree observations outperform Screenshot. However, subtask success rates are substantially higher than task success rates in every setting. For example, Claude Opus 4.6 under Portal Guidance + Accessibility Tree reaches 89.7\% subtask success but only 51.9\% task success, while Claude Opus 4.6 CUA reaches 78.4\% subtask success versus 36.3\% task success. This gap supports the observation in Section~\ref{sec:discussion} that agents can execute individual steps correctly but often fail to complete full workflows end-to-end.

\subsection{Head-to-Head Agent Comparisons}
In the main results, we report overall task and subtask success rates with confidence intervals for each agent. In many settings, however, the more relevant question is whether one agent significantly outperforms another. We therefore report head-to-head comparisons based on pairwise differences in performance, together with confidence intervals for those differences. These comparisons show that \textsc{HealthAdminBench} provides enough signal to detect many statistically significant differences between agents. Table~\ref{tab:head_to_head_task} reports head-to-head comparisons for task success rate, and Table~\ref{tab:head_to_head_subtask} reports the corresponding comparisons for subtask success rate.

\begin{table}[ht]
\centering
\scriptsize
\renewcommand{\arraystretch}{1.25}
\setlength{\tabcolsep}{6pt}
\begin{tabular}{lccccccc}
\hline
\makecell[l]{Baseline\\Compare} & \makecell{Claude\\Opus 4.6 CUA} & \makecell{GPT-5.4\\CUA} & \makecell{Kimi K2.5} & \makecell{Claude\\Opus 4.6} & \makecell{Qwen 3.5} & \makecell{Gemini\\3.1 Pro} \\
\hline
\makecell[l]{GPT-5.4\\CUA} & \cellcolor{green!32!white}\makecell{\textbf{9.6\%}\\\textbf{(2.2\% - 17.8\%)}} &  &  &  &  &  \\
\makecell[l]{Kimi K2.5} & \cellcolor{green!68!white}\makecell{\textbf{20.7\%}\\\textbf{(11.8\% - 29.6\%)}} & \cellcolor{green!37!white}\makecell{\textbf{11.1\%}\\\textbf{(3.7\% - 18.5\%)}} &  &  &  & \\
\makecell[l]{Claude\\Opus 4.6} & \cellcolor{green!71!white}\makecell{\textbf{21.5\%}\\\textbf{(13.3\% - 29.6\%)}} & \cellcolor{green!39!white}\makecell{\textbf{11.9\%}\\\textbf{(3.0\% - 20.7\%)}} & \cellcolor{green!2!white}\makecell{0.7\%\\(-5.2\% - 7.4\%)} &  &  &  \\
\makecell[l]{Qwen 3.5} & \cellcolor{green!76!white}\makecell{\textbf{23.0\%}\\\textbf{(14.1\% - 31.1\%)}} & \cellcolor{green!44!white}\makecell{\textbf{13.3\%}\\\textbf{(5.9\% - 20.7\%)}} & \cellcolor{green!7!white}\makecell{2.2\%\\(-4.4\% - 8.9\%)} & \cellcolor{green!5!white}\makecell{1.5\%\\(-5.2\% - 8.2\%)} &  &  \\
\makecell[l]{Gemini\\3.1 Pro} & \cellcolor{green!80!white}\makecell{\textbf{24.4\%}\\\textbf{(16.3\% - 33.3\%)}} & \cellcolor{green!49!white}\makecell{\textbf{14.8\%}\\\textbf{(7.4\% - 23.0\%)}} & \cellcolor{green!12!white}\makecell{3.7\%\\(-3.7\% - 11.1\%)} & \cellcolor{green!10!white}\makecell{3.0\%\\(-4.4\% - 11.1\%)} & \cellcolor{green!5!white}\makecell{1.5\%\\(-5.2\% - 8.1\%)} &  \\
\makecell[l]{GPT-5.4} & \cellcolor{green!100!white}\makecell{\textbf{30.4\%}\\\textbf{(22.2\% - 39.3\%)}} & \cellcolor{green!68!white}\makecell{\textbf{20.7\%}\\\textbf{(13.3\% - 28.9\%)}} & \cellcolor{green!32!white}\makecell{\textbf{9.6\%}\\\textbf{(3.0\% - 17.0\%)}} & \cellcolor{green!29!white}\makecell{\textbf{8.9\%}\\\textbf{(2.2\% - 15.6\%)}} & \cellcolor{green!24!white}\makecell{\textbf{7.4\%}\\\textbf{(0.7\% - 13.3\%)}} & \cellcolor{green!20!white}\makecell{5.9\%\\(0.0\% - 11.9\%)} \\
\hline
\end{tabular}
\caption{Head-to-head differences in task success rate (percentage points). Positive numbers indicate that the agent in the column is better than the agent in row. 95\% test-set bootstrap confidence intervals are reported under the overall difference between agents. Bolded rows are statistically significant ($p < 0.05$).}
\label{tab:head_to_head_task}
\end{table}

\begin{table}[ht]
\centering
\scriptsize
\renewcommand{\arraystretch}{1.25}
\setlength{\tabcolsep}{6pt}
\begin{tabular}{lcccccc}
\hline
\makecell[l]{Baseline\\Compare} & \makecell{Claude\\Opus 4.6 CUA} & \makecell{GPT-5.4\\CUA} & \makecell{Kimi K2.5} & \makecell{Claude\\Opus 4.6} & \makecell{Qwen 3.5} & \makecell{Gemini\\3.1 Pro} \\
\hline
\makecell[l]{GPT-5.4\\CUA} & \cellcolor{red!12!white}\makecell{\textbf{-4.4\%}\\\textbf{(-8.8\% - -0.3\%)}} &  &  &  &  &  \\
\makecell[l]{Kimi K2.5} & \cellcolor{green!56!white}\makecell{\textbf{20.9\%}\\\textbf{(14.5\% - 27.1\%)}} & \cellcolor{green!68!white}\makecell{\textbf{25.3\%}\\\textbf{(19.9\% - 30.7\%)}} &  &  &  &  \\
\makecell[l]{Claude\\Opus 4.6} & \cellcolor{green!62!white}\makecell{\textbf{23.0\%}\\\textbf{(15.4\% - 30.5\%)}} & \cellcolor{green!74!white}\makecell{\textbf{27.4\%}\\\textbf{(21.0\% - 33.7\%)}} & \cellcolor{green!6!white}\makecell{2.1\%\\(-4.9\% - 9.4\%)} &  &  &  \\
\makecell[l]{Qwen 3.5} & \cellcolor{green!44!white}\makecell{\textbf{16.4\%}\\\textbf{(9.9\% - 22.9\%)}} & \cellcolor{green!56!white}\makecell{\textbf{20.8\%}\\\textbf{(15.7\% - 25.9\%)}} & \cellcolor{red!12!white}\makecell{-4.5\%\\(-10.8\% - 1.7\%)} & \cellcolor{red!18!white}\makecell{-6.6\%\\(-13.4\% - 0.1\%)} &  &  \\
\makecell[l]{Gemini\\3.1 Pro} & \cellcolor{green!19!white}\makecell{\textbf{6.9\%}\\\textbf{(2.3\% - 11.4\%)}} & \cellcolor{green!31!white}\makecell{\textbf{11.4\%}\\\textbf{(8.2\% - 14.6\%)}} & \cellcolor{red!37!white}\makecell{\textbf{-13.9\%}\\\textbf{(-19.4\% - -8.4\%)}} & \cellcolor{red!43!white}\makecell{\textbf{-16.0\%}\\\textbf{(-23.1\% - -8.9\%)}} & \cellcolor{red!25!white}\makecell{\textbf{-9.5\%}\\\textbf{(-15.2\% - -4.1\%)}} &  \\
\makecell[l]{GPT-5.4} & \cellcolor{green!88!white}\makecell{\textbf{32.8\%}\\\textbf{(26.6\% - 38.9\%)}} & \cellcolor{green!100!white}\makecell{\textbf{37.2\%}\\\textbf{(31.9\% - 42.5\%)}} & \cellcolor{green!32!white}\makecell{\textbf{11.9\%}\\\textbf{(5.6\% - 18.3\%)}} & \cellcolor{green!26!white}\makecell{\textbf{9.8\%}\\\textbf{(2.5\% - 17.2\%)}} & \cellcolor{green!44!white}\makecell{\textbf{16.4\%}\\\textbf{(10.8\% - 22.1\%)}} & \cellcolor{green!69!white}\makecell{\textbf{25.9\%}\\\textbf{(19.9\% - 31.7\%)}} \\
\hline
\end{tabular}
\caption{Head-to-head differences in subtask success rate (percentage points). Positive numbers indicate that the agent in the column is better than the agent in row. 95\% test-set bootstrap confidence intervals are reported under the overall difference between agents. Bolded rows are statistically significant ($p < 0.05$).}
\label{tab:head_to_head_subtask}
\end{table}
\section{Fine-Tuning on Domain-Specific Data}
\label{sec:finetune}

\paragraph{Experimental Setup.}
All fine-tuning experiments are performed with accessibility-tree observations and Task Description prompting to reduce cost and complexity. We split \textsc{HealthAdminBench} tasks into disjoint training and test sets by randomly sampling 35 tasks for our test set. For each of the 100 remaining training tasks, we generate trajectories using the task-specific step-by-step prompting setting described in Section~\ref{section:promts}, which explicitly specifies the intended workflow steps for the task. These trajectories are then used as supervision to perform LoRA fine-tuning on top of the base Qwen-3.5-27B model, resulting in Qwen-3.5-Kinetic-SFT. We use the Tinker API for this experiment. We train for a single epoch with a learning rate of 3.4e-4, a LoRA rank of 8, a batch size of 16, and a max sequence length of 24,574.

\paragraph{Evaluation Protocol.}
The fine-tuned Qwen-3.5-Kinetic-SFT model is evaluated on the held-out test set of 35 examples. We evaluate Claude Opus 4.6 on the same held-out test set of 35 examples, without further fine-tuning.

\paragraph{Results.}

As shown in Table \ref{tab:finetuning}, Qwen-3.5-Kinetic-SFT achieves a task success rate of 40\%, representing a +23\% absolute improvement over the base Qwen 3.5 27B model. This represents a significant gain over the out-of-the-box performance of the best performing frontier model, Claude Opus 4.6, which achieved only a 25.7\% success rate on the 35 held out tasks.

\begin{table}[h]
\centering
\small
\caption{Results of domain-specific finetuning on 35 held-out examples using accessibility tree observations and task description + portal guidance prompting.}
\label{tab:finetuning}
\setlength{\tabcolsep}{6pt}
\begin{tabular}{lrrr}
\toprule
\textbf{Model} & \textbf{Task Success Rate} &  \textbf{Subtask Success Rate} & \textbf{Fine Tuned?}  \\
\midrule
Claude Opus 4.6 & 25.7\% & 75.0\% &  - \\
Qwen-3.5 & 17.1\% & 65.4\% & - \\
\midrule
Qwen-3.5-Kinetic-SFT & \textbf{40.0\%} & \textbf{83.3\%} &  \checkmark \\
\bottomrule
\end{tabular}
\end{table}

This experiment is performed on a very small test set, so it is necessary to be very careful about statistical noise. To address this, we perform head-to-head bootstrap comparisons in Tables \ref{tab:finetuning_head_to_head_task} and \ref{tab:finetuning_head_to_head_subtask}.  We find that increase in subtask performance of Qwen-3.5-Kinetic-SFT compared to Claude Opus 4.6 is statistically significant, but that the task performance is not.

\begin{table}[ht]
\centering
\scriptsize
\renewcommand{\arraystretch}{1.25}
\setlength{\tabcolsep}{6pt}
\begin{tabular}{lccc}
\hline
\makecell[l]{Baseline\\Compare} & \makecell{Qwen\\-3.5-Kinetic-SFT} & \makecell{Claude\\Opus 4.6} & \makecell{Qwen-3.5} \\
\hline
\makecell[l]{Claude\\Opus 4.6} & \cellcolor{green!62!white}\makecell{14.3\%\\(-2.9\% - 31.4\%)} &  &  \\
\makecell[l]{Qwen-3.5} & \cellcolor{green!100!white}\makecell{\textbf{22.9\%}\\\textbf{(8.6\% - 37.1\%)}} & \cellcolor{green!38!white}\makecell{8.6\%\\(-2.9\% - 20.0\%)} &  \\
\hline
\end{tabular}
\caption{Head-to-head differences in task success rate for the 35-task finetuning comparison (percentage points).}
\label{tab:finetuning_head_to_head_task}
\end{table}

\begin{table}[ht]
\centering
\scriptsize
\renewcommand{\arraystretch}{1.25}
\setlength{\tabcolsep}{6pt}
\begin{tabular}{lccc}
\hline
\makecell[l]{Baseline\\Compare} & \makecell{Qwen\\-3.5-Kinetic-SFT} & \makecell{Claude\\Opus 4.6} & \makecell{Qwen-3.5} \\
\hline
\makecell[l]{Claude\\Opus 4.6} & \cellcolor{green!47!white}\makecell{\textbf{8.4\%}\\\textbf{(1.2\% - 15.4\%)}} &  &  \\
\makecell[l]{Qwen-3.5} & \cellcolor{green!100!white}\makecell{\textbf{17.9\%}\\\textbf{(7.2\% - 29.8\%)}} & \cellcolor{green!53!white}\makecell{9.5\%\\(-1.2\% - 20.1\%)} &  \\
\hline
\end{tabular}
\caption{Head-to-head differences in subtask success rate for the 35-task finetuning comparison (percentage points).}
\label{tab:finetuning_head_to_head_subtask}
\end{table}

These results suggest that even limited amounts of domain-specific supervision -- in our case, only 100 examples -- can substantially improve performance on complex administrative workflows, highlighting the potential for rapid adaptation in enterprise settings and the importance of domain-specific workflow data.

\paragraph{Limitations.}
This experiment is conducted in only one benchmark configuration (Task Description prompting and accessibility-tree observations) on a relatively small subset of tasks, and results may vary with different splits or larger-scale fine-tuning. 


\end{document}